# NIT_COVID-19 at WNUT-2020 Task 2: Deep Learning Model RoBERTa for Identify Informative COVID-19 English Tweets

**Jagadeesh M S, Alphonse P J A**
Department of Computer Applications
National Institute of Technology
Tiruchirapalli,TamilNadu, India
`malla.sree@gmail.com`[1],
`alphonse@nitt.edu`[2]

## Abstract

This paper presents the model submitted by NIT_COVID-19 team for identified informative COVID-19 English tweets at WNUT-2020 Task2. This shared task addresses the problem of automatically identifying whether an English tweet related to informative (novel coronavirus) or not. These informative tweets provide information about recovered, confirmed, suspected, and death cases as well as location or travel history of the cases. The proposed approach includes pre-processing techniques and pre-trained RoBERTa with suitable hyperparameters for English coronavirus tweet classification. The performance achieved by the proposed model for shared task WNUT 2020 Task2 is 89.14% in the F1-score metric.

## 1 Introduction

Present the world is suffering from a novel coronavirus (Linton et al., 2020) from December-2019 to till date. It spreads all the continents and almost all countries in the world. Now a days many online tools or resources (Xu et al., 2020) provide coronavirus information to people in the world, but these resources are not continuously up to date. They are updating in a particular time interval. The world's major trusted resources like WHO (World Health Organization) (Chakraborty and Maity, 2020) also update this COVID-19 related information once a day. This pandemic COVID-19 has been spreading rapidly(Shao et al., 2020). In this situation world is looking for an automatic monitoring system for identifying useful information about COVID-19.

One of the most popular and trusted online social network platform (Karampelas, 2013) is TWITTER, an alternative source for updating the present pandemic information. Twitter is getting nearly 4 million COVID-19 tweets daily (Lopez et al., 2020), but only a few of them are informative. Twitter API provides tweet information openly to the research community. This is very useful to identify automatically informative COVID-19 tweets.

In this WNUT-2020 shared task-2, we proposed a model for automatically identifying COVID-19 informative tweets, which describes information about confirmed, recovered, and suspected and death cases as well as location or travel history of the COVID-19 affected patients. We have used pre-trained deep learning "RoBERTa" (Liu et al., 2019) with word-level embedding along with some pre-processing techniques. We also experimented with CNN (Moriya and Shibata, 2018) and pre-trained BERT (Devlin et al., 2018)– encoded sentences as input. The proposed deep learning Model gives a high F1- score compares to the above approaches. After discussed the related work in section 2, section 3 discussed the methodology and the data in detail and discusses the results in section 4. We analyze the results and error analysis in section 5 and section 6 concludes the paper.

## 2 Realated Work

Informative tweet identification has been of interest for researchers in recent years. Early work in the related fields include detection of online media (ALRashdi and O'Keefe, 2019), racism (Kabir and Madria, 2019), and disaster (Sreenivasulu and Sridevi, 2020). Papers published in recent years include (Madichetty and Sridevi, 2019), which introduces the tweet classification detection dataset and experiments with different machine learning models, such as naıve Bayes, logistic regression, random forests, and linear SVMs to investigate hate speech and disrespectful language, which experiments further on the same dataset using SVMs with n-grams and skip-grams features, and (Gambäck and Sikdar, 2017) and (Bohra et al., 2018), both exploring the performance of neural networks and comparing them with other machine learning ap-proaches. Also, there has been published a couple of surveys covering various work addressing the identification of abusive, toxic, and offensive language, hate speech, etc., and their methodology including (Schmidt and Wiegand, 2017) and (Fortuna and Nunes, 2018). Additionally, there were several workshops and shared tasks on offensive language identification and related problems, including TA-COS2, Abusive Language Online3, and TRAC4 and GermEval (Wiegand et al., 2018), which shows the significance of the problem.

## 3 Methodology

The methodology used for WNUT-2020 Task 2, consists of a preprocessing phase and a deep learning model implementation phase.

### 3.1 Pre-processing

This phase consists of

1. Tokenization: In this step, the entire sentence (Pitsilis et al., 2018)is split into words (Tokens). Python "nltk" package helps to split the tweet into the tokens.

2. Convert Tokens to lower cases: The tokens from tokenization are may in lower or upper cases. In this step convert all tokens convert into the same case(here lower case).

3. Filter out punctuation: Remove all punctuation's (Gupta and Joshi, 2017) from the tokens.

4. Filter out stop words(and pipeline): Remove commonly used English words (Munková et al., 2013)( English stop words)

5. Stem words: Simply convert tokens into their genitive singular form (Grefenstette, 1996). For example, the wait is a stem word for waiting and waited.

### 3.2 Deep Learning Model

The goal of this WNUT-2020 task 2 is to identify a given COVID-19 tweet that is INFORMATIVE or UNINFORMATIVE. In this task, we have used a pre-trained deep learning model RoBERTa (Robustly Optimized BERT Approach), which is an optimized model for BERT (Jawahar et al., 2019). RoBERTa has features like

1. Train the data up to 160 GB.

2. Increase the number of iterations up to 500k.

3. Train the model with batch size 8k.

4. Larger byte-level BPE vocabulary with 50k sub word units.

5. Dynamically changing the masking pattern applied to the training data.

We trained the RoBERTa model with different combinations of hyperparameters for the given dataset, which was provided by WNUT-2020. Finally, we have gotten better metrics for hyper parametric values.

- Used 'roberta-base'
- Maximum learning rate is equal to 1e-5.
- We used batch size is equal to 16.
- Maximum sequence length of tweet in the dataset is 143.
- Avoid over-fitting we set hidden dropout is equal to 0.05.
- Hidden size for 'roberta-base' is equal to 768.
- An 'adam' is used for optimizer.
- Trained for 50 epochs.

### 3.3 Baseline Methods

we used three baseline methods:

- An Random Forest (Bhagat and Patil, 2015) with maximum depth of 26 and no of estimators 500.

- An SVM (Penagarikano et al., 2011) with 1- to 3-gram word TFIDF character count feature vectors as input.

- An CNN (Zhang et al., 2018) with embedding layer output size of 128 and fully connected. "adam" is used as an optimizer.

The CNN was trained for 25 epochs with stochastic gradient descent. Using Scikit-learn (Pedregosa et al., 2011), the baseline methods were implemented.

| Data Set | INFORMATIVE | UNINFORMATIVE |
|---|---|---|
| Training | 3303 | 3697 |
| Validation | 472 | 528 |
| Test | 944 | 1056 |

Table 1: COVID-19 English Tweets Data Set Details

## 4 Data

The main dataset used to train our model is COVID-19 INFORMATIVE English tweets identification (COVID-19 tweet dataset)(Nguyen et al., 2020), which was provided by WNUT-2020 at Task 2. This dataset consists of training, validation, and testing phases with 7000, 1000, and 12000 tweet records respectively. Training and validation tweet records have fields name with Id, Text, and Label. The tweet classification was done, based on the tweet information, which includes recovered, suspected, confirmed death cases, as well as location or travel history of the cases that come under INFORMATIVE tweets and other non-informative COVID-19 tweets, comes under UNINFORMATIVE.

The 'COVID-19 English Tweets' dataset provides 3303 INFORMATIVE tweets and 3697 UNINFORMATIVE tweets from the training dataset, 472 INFORMATIVE, and 528 UNINFORMATIVE tweets from the validation dataset. The test dataset is a large set of 12k tweets and the actual 2k test tweets test the model in the form of an F1-Score metric. This test set consists of 944 INFORMATIVE and 1056 UNINFORMATIVE unlabeled tweets as shown in table 1.

## 5 Results and Error Analysis

Finally, We done experiment on the baseline models in 3.3 and the model described in section 3.2 using the COVID-19 English tweets data sets. The task organizers did not provide any baseline scores for this task. For better results on the validation data set, RoBERTa model trained on training data set, tested on validation data set and obtained results shown in the table 2. Models of deep learning need a large amount of data for training. So for better results on the test data set, RoBERTa model trained on both training data set as well as validation data set and tested on test data set and obtained results shown in the table 3. The best results are noted in bold. This data comprises tweets with INFORMATIVE and UNINFORMATIVE labels in binary classification. RoBERTa gives best performance on validation data by a margin of more than 0.2 with the best baseline performance(BERT) (Karisani and Karisani, 2020). From the confusing matrix as shown in Figure 1 we observe that the performance of RoBERTa on INFORMATIVE is quite good, but not on UNINFORMATIVE. we can observe the detailed results of RoBERTa in table 4.

| Model | F1-score | Accuracy |
|---|---|---|
| Random Forest | 81.5894 | 82.0688 |
| SVM | 82.7105 | 82.0488 |
| CNN | 83.7370 | 83.0691 |
| BERT | 88.9787 | 89.1006 |
| **RoBERTa** | **89.1864** | **89.5000** |

Table 2: Results obtained on the Validation Set.

| Model | F1-score | Accuracy | Recall | precision |
|---|---|---|---|---|
| **RoBERTa** | **89.14** | **89.35** | **92.58** | **85.94** |

Table 3: Results obtained on the Test Data Set.

We have submitted results of our model on test data as part of WNUT-2020 Task 2 for recognising COVID-19 English informative tweets. The organizers released unlabeled test data for the final evaluation phase. This is a large set of 12K tweets, and the actual 2K test tweets by which our model has evaluated are hidden in this large set. Our model output metric values on test data are F1-score is 89.14, precision is 85.94, recall is 92.35, and accuracy value is 89.35 as shown in table 3.

RoBERTa outperformed well,compare with BERT, by 0.20 % F1-score. The baseline model results, however,worse than the RoBERTa and , probably due the fact that the RoBERTa model requires fine-tuning for more task specific representations. The majority of RoBERTa errors are in INFORMATIVE class.

## 6 Conclusion and Future Work

In this (WNUT-2020 shared Task2) competition, We introduced the NIT COVID-19 team's ap-

| Label | F1-score | Accuracy |
|---|---|---|
| INFORMATIVE | 89.89 | 90.23 |
| UNINFORMATIVE | 88.16 | 88.32 |

Table 4: Detailed RoBERTa Model Metric values on validation data set

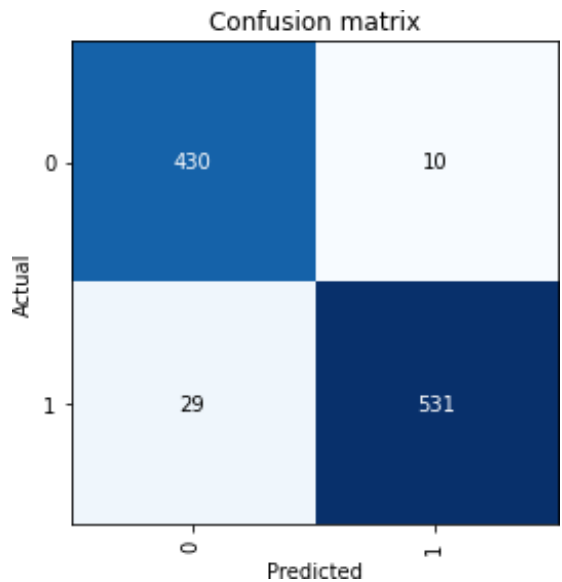

0 - Represents INFORMATIVE
1 - Represents UNINFORMATIVE

Figure 1: Confusion Matrix for validation set

proach to the issue of informative tweet recognition and automatic categorization from the COVID-19 tweet master dataset of informative tweets. The pre-trained deep learning RoBERTa model outperformed the remaining models, including Random Forest, SVM, CNN and BERT. Furthermore, the analysis of the results indicates the some of INFORMATIVE tweets are not identified by the model. Such deficiencies demand larger training corpa and need a prominent features for training. In future work we will concentrate on latest pre-trained deep learning models and other issues for better results.

## References

Reem ALRashdi and Simon O'Keefe. 2019. Deep learning and word embeddings for tweet classification for crisis response. *arXiv preprint arXiv:1903.11024*.

Reshma C Bhagat and Sachin S Patil. 2015. Enhanced smote algorithm for classification of imbalanced big-data using random forest. In *2015 IEEE International Advance Computing Conference (IACC)*, pages 403–408. IEEE.

Aditya Bohra, Deepanshu Vijay, Vinay Singh, Syed Sarfaraz Akhtar, and Manish Shrivastava. 2018. A dataset of hindi-english code-mixed social media text for hate speech detection. In *Proceedings of the second workshop on computational modeling of people's opinions, personality, and emotions in social media*, pages 36–41.

Indranil Chakraborty and Prasenjit Maity. 2020. Covid-19 outbreak: Migration, effects on society, global environment and prevention. *Science of the Total Environment*, page 138882.

Jacob Devlin, Ming-Wei Chang, Kenton Lee, and Kristina Toutanova. 2018. Bert: Pre-training of deep bidirectional transformers for language understanding. *arXiv preprint arXiv:1810.04805*.

Paula Fortuna and Sérgio Nunes. 2018. A survey on automatic detection of hate speech in text. *ACM Computing Surveys (CSUR)*, 51(4):1–30.

Björn Gambäck and Utpal Kumar Sikdar. 2017. Using convolutional neural networks to classify hate-speech. In *Proceedings of the first workshop on abusive language online*, pages 85–90.

David A Hull Gregory Grefenstette. 1996. A detailed analysis of english stemming algorithms. *Rank Xerox Research Centre*, 6:1–16.

Itisha Gupta and Nisheeth Joshi. 2017. Tweet normalization: A knowledge based approach. In *2017 International Conference on Infocom Technologies and Unmanned Systems (Trends and Future Directions)(ICTUS)*, pages 157–162. IEEE.

Ganesh Jawahar, Benoît Sagot, and Djamé Seddah. 2019. What does bert learn about the structure of language?

Md Yasin Kabir and Sanjay Madria. 2019. A deep learning approach for tweet classification and rescue scheduling for effective disaster management. In *Proceedings of the 27th ACM SIGSPATIAL International Conference on Advances in Geographic Information Systems*, pages 269–278.

Panagiotis Karampelas. 2013. *Techniques and tools for designing an online social network platform.* Springer.

Negin Karisani and Payam Karisani. 2020. Mining coronavirus (covid-19) posts in social media. *arXiv preprint arXiv:2004.06778*.

Natalie M Linton, Tetsuro Kobayashi, Yichi Yang, Katsuma Hayashi, Andrei R Akhmetzhanov, Sung-mok Jung, Baoyin Yuan, Ryo Kinoshita, and Hiroshi Nishiura. 2020. Incubation period and other epidemiological characteristics of 2019 novel coronavirus infections with right truncation: a statistical analysis of publicly available case data. *Journal of clinical medicine*, 9(2):538.

Yinhan Liu, Myle Ott, Naman Goyal, Jingfei Du, Mandar Joshi, Danqi Chen, Omer Levy, Mike Lewis, Luke Zettlemoyer, and Veselin Stoyanov. 2019. Roberta: A robustly optimized bert pretraining approach. *arXiv preprint arXiv:1907.11692*.

Christian E Lopez, Malolan Vasu, and Caleb Gallemore. 2020. Understanding the perception of covid-19 policies by mining a multilanguage twitter dataset. *arXiv preprint arXiv:2003.10359*.

Sreenivasulu Madichetty and M Sridevi. 2019. Detecting informative tweets during disaster using deep neural networks. In *2019 11th International Conference on Communication Systems & Networks (COMSNETS)*, pages 709–713. IEEE.

Shun Moriya and Chihiro Shibata. 2018. Transfer learning method for very deep cnn for text classification and methods for its evaluation. In *2018 IEEE 42nd Annual Computer Software and Applications Conference (COMPSAC)*, volume 2, pages 153–158. IEEE.

Daša Munková, Michal Munk, and Martin Vozár. 2013. Influence of stop-words removal on sequence patterns identification within comparable corpora. In *International Conference on ICT Innovations*, pages 67–76. Springer.

Dat Quoc Nguyen, Thanh Vu, Afshin Rahimi, Mai Hoang Dao, Linh The Nguyen, and Long Doan. 2020. WNUT-2020 Task 2: Identification of Informative COVID-19 English Tweets. In *Proceedings of the 6th Workshop on Noisy User-generated Text*.

F. Pedregosa, G. Varoquaux, A. Gramfort, V. Michel, B. Thirion, O. Grisel, M. Blondel, P. Prettenhofer, R. Weiss, V. Dubourg, J. Vanderplas, A. Passos, D. Cournapeau, M. Brucher, M. Perrot, and E. Duchesnay. 2011. Scikit-learn: Machine learning in Python. *Journal of Machine Learning Research*, 12:2825–2830.

Mikel Penagarikano, Amparo Varona, Luis Javier Rodriguez-Fuentes, and German Bordel. 2011. Dimensionality reduction for using high-order n-grams in svm-based phonotactic language recognition. In *Twelfth Annual Conference of the International Speech Communication Association*.

Georgios K Pitsilis, Heri Ramampiaro, and Helge Langseth. 2018. Detecting offensive language in tweets using deep learning. *arXiv preprint arXiv:1801.04433*.

Anna Schmidt and Michael Wiegand. 2017. A survey on hate speech detection using natural language processing. In *Proceedings of the Fifth International workshop on natural language processing for social media*, pages 1–10.

Nian Shao, Yu Chen, Jin Cheng, and Wen Bin Chen. 2020. Some novel statistical time delay dynamic model by statistics data from ccdc on novel coronavirus pneumonia. *Kongzhi Lilun Yu Yingyong/Control Theory and Applications*, 37(4).

Madichetty Sreenivasulu and M Sridevi. 2020. Comparative study of statistical features to detect the target event during disaster. *Big Data Mining and Analytics*, 3(2):121–130.

Michael Wiegand, Melanie Siegel, and Josef Ruppenhofer. 2018. Overview of the germeval 2018 shared task on the identification of offensive language.

Bo Xu, Moritz UG Kraemer, Bernardo Gutierrez, Sumiko Mekaru, Kara Sewalk, Alyssa Loskill, Lin Wang, Emily Cohn, Sarah Hill, Alexander Zarebski, et al. 2020. Open access epidemiological data from the covid-19 outbreak. *The Lancet Infectious Diseases*, 20(5):534.

Jiarui Zhang, Yingxiang Li, Juan Tian, and Tongyan Li. 2018. Lstm-cnn hybrid model for text classification. In *2018 IEEE 3rd Advanced Information Technology, Electronic and Automation Control Conference (IAEAC)*, pages 1675–1680. IEEE.